\RequirePackage{fix-cm}

\documentclass[twocolumn, natbib]{svjour3}

\smartqed  

\usepackage[ngerman,english]{babel}

\usepackage{graphicx}

\usepackage{tikz}
\usepackage{pgfplots}
\usepackage{pgfplotstable}
\usepackage{xint}
\usepackage{xcolor}
\usepackage{mathtools} 
\usepackage{url} 

\usepackage{booktabs}
\usetikzlibrary{pgfplots.colormaps}
\pgfplotsset{compat=1.16}
\usepackage{pgfplotstable}
\def\moduloop#1#2{\ifnum\numexpr(#1-(#1/#2)*(#2))\relax<0(#1-(#1/#2)*(#2)+#2)\else(#1-(#1/#2)*(#2))\fi}

\journalname{PFG -- Journal of Photogrammetry, Remote Sensing and Geoinformation Science}

\definecolor{revision}{RGB}{0,0,0}

\begin{document}

\title{Urban Change Forecasting from Satellite Images}

\author{Nando~Metzger         \and 
        Mehmet~Özgür~Türkoglu \and
        Rodrigo~Caye~Daudt \and
        Jan~Dirk~Wegner \and
        Konrad~Schindler
}

\institute{Nando Metzger \at
            Photogrammetry and Remote Sensing, ETH Zürich,
            \email{nando.metzger@geod.baug.ethz.ch}, 
            \url{orcid.org/0000-0002-0299-3064}
            \and
            Mehmet~Özgür~Türkoglu \at Photogrammetry and Remote Sensing, ETH Zürich 
            \url{orcid.org/0000-0003-1446-2778}
           \and
           Rodrigo~Caye~Daudt \at Photogrammetry and Remote Sensing, ETH Zürich,
           \url{orcid.org/0000-0002-4952-9736}
           \and
           Jan~Dirk~Wegner \at Institute for Computational Science, University of Zurich
           \url{orcid.org/0000-0002-0290-6901}
           \and
           Konrad Schindler \at Photogrammetry and Remote Sensing, ETH Zürich,
           \url{orcid.org/0000-0002-3172-9246} 
}

\date{Received: date / Accepted: date}

\maketitle

\begin{abstract}
Forecasting where and when new buildings will emerge is a rather unexplored topic, but one that is very useful in many disciplines such as urban planning, agriculture, resource management, and even autonomous \textcolor{revision}{flying}.
In the present work, we present a method that accomplishes this task with a deep neural network and a custom pretraining procedure.
In \emph{Stage~1}, a U-Net backbone is pretrained within a Siamese network architecture that aims to solve a (building) change detection task.
In \emph{Stage~2}, the backbone is repurposed to forecast the emergence of new buildings based solely on one image acquired before its construction.
Furthermore, we also present a model that forecasts the time range within which the change will occur.
We validate our approach using the SpaceNet7 dataset, \textcolor{revision}{which} covers an area of 960 km$^2$ at 24 points in time across two years.
In our experiments, we found that our proposed pretraining method consistently outperforms the traditional pretraining using the ImageNet dataset.
We also show that it is to some degree possible to predict in advance when building changes will occur.

\end{abstract}


\section{Introduction}
\label{intro}

Understanding the evolution of land use (e.g., constructions and changes of the land cover) is crucial in fields like urban planning, agriculture, natural resource management, anticipating housing market prices, and even autonomous driving and flying.
For the latter, visual positioning systems that do not rely on GNSS are a concrete application example \citep{daedalean2021VPS}. One of the critical components in a purely vision-based positioning system is an up-to-date map of the environment.
Especially in emergencies, it is crucial to have a map with maximal confidence to find safe landing sites, and hence, regular map updating missions must be conducted, often involving resource-intensive survey flights.
A system that is able to anticipate the changes could assist in indicating the locations of future survey flights and potentially save unnecessary flights in regions with no change.

\textcolor{revision}{Predicting urban transformations is a complex endeavor that requires advanced image understanding. Nonetheless, current research predominantly employs traditional non-data-driven methods or relies on pixel-wise Multi-layer perceptrons (MLPs). In response, we strive to bridge this research void by introducing a data-centric methodology, which enables the effective training of fully-convolutional neural networks specifically designed to anticipate urban change. }

\begin{figure*}[htbp]
    \centering
    \includegraphics[width=\textwidth]{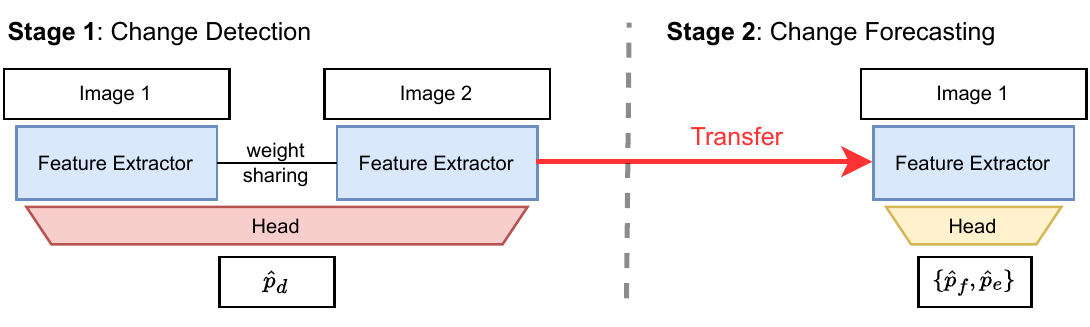}
    \caption{Modular model architectures and general workflow. In Stage 1, 
    we train a feature extractor \textcolor{revision}{(a U-Net with ResNet50 encoder) in a Siamese setup to detect the pixel-wise urban changes $\hat{p}_d$ from two satellite images. In stage 2,} the feature extractor is repurposed to \textcolor{revision}{be used} for the change forecasting task, which is to forecast the change as $\hat{p}_f$ and also predict when it will happen as $\hat{p}_e$. \textcolor{revision}{The ``Head"-CNN is not transferred, but trained separately for each stage.}}
    \label{fig:graphical_abstract}
\end{figure*}

In the present work, we aim to forecast where and when changes in building footprints are going to happen.
Change forecasting is a binary segmentation problem with labels ``change" and ``no change", where the forecasting range \textcolor{revision}{(i.e., the time span between the acquisition of the query image and the actual change)} is defined implicitly, \textcolor{revision}{by selecting training samples with a fixed forecasting range.}
We use satellite images as the primary input data source because they provide global, uniform coverage. 
The SpaceNet7 dataset provides an adequate compromise that balances our requirements. The publicly available, \textcolor{revision}{annotated subset} comprises 60 locations with a total extent of 960 km$^{2}$ and a ground sampling distance (GSD) of 4 m.   
The dataset consists of up to 24 image time-steps per location, with a temporal resolution of 1 month, which allows us to analyze the model's performances with respect to different time-range.
We develop a 2-stage training strategy (as shown in Figure~\ref{fig:graphical_abstract}) that is centered around a U-Net segmentation backbone \citep{ronneberger2015u}.

\begin{itemize}
    \item In \emph{Stage~1}, we train a change detection network with a Siamese network layout based on a U-Net backbone, using as input pairs of satellite images of the same location, with different time stamps.
    \item In \emph{Stage~2}, we keep the backbone of \emph{Stage~1} and use it as a feature extractor for change forecasting. Moreover, we slightly adapt the architecture to also produce time range forecasts, \textcolor{revision}{in which} the model needs to anticipate \textcolor{revision}{within} which time window the change will occur. We implement this task via a multi-task learning setup, where the model needs to predict an ordinal label \textcolor{revision}{that indicates when a change is expected to} happen, in addition to the binary change label.
\end{itemize}

\textcolor{revision}{
\section{Related Work} \label{sec:related}
}

The exploration of change detection is a well-explored task in remote sensing. Early approaches relied on hand-crafted workflows to detect changes in satellite images, while later, classical machine learning approaches were used to automatically classify the hand-crafted features~\citep{singh1989review,hussain2013change,le2013urban,metzlaff2015region,wessels2016rapid}. With the rise of deep learning techniques, \textcolor{revision}{researchers} started to employ neural networks that are capable of learning features themselves~\citep{el2017zoom,zhu2017change,liu2020building}.
\textcolor{revision}{Siamese Networks that share the same feature extractor across multiple images turned out to be a suitable inductive bias for several computer vision tasks, including stereo matching~\citep{zagoruyko2015learning} and object tracking~\citep{bertinetto2016fully}. But also in remote sensing, researchers employed Siamese feature extractors~\citep{zhan2017change}, often as part of an end-to-end trainable network~\citep{daudt2018urban,daudt2018fully,arabi2018optical}. Recent studies have extended this concept to multi-task scenarios~\citep{liu2019siamese}, network architectures based on attention mechanisms~\citep{chen2020dasnet} and multi-scale features~\citep{yang2021deep}.
}

There have been a few attempts to predict future land cover with traditional machine learning.
\citet{iacono2012markov} make the rigid assumption that the land use/land cover (LULC) class is a discrete state depending solely on its previous state. In this way, they are able to apply Markov chains to model state changes over time. However, this relatively strong assumption may not always be valid and limits the use of auxiliary data.
Land transformation models \citep{pijanowski2002using,tang2005forecasting,newman2016using,pijanowski2020land} are established methods for LULC class forecasting that allow the inclusion of additional social, political, and environmental drivers and process them via an MLP.
\citet{chu2010forecasting} used Markov Chains (MC) to forecast land use changes, while later, \citet{nguyen2020land} proposed an approach that employs satellite imagery as a driving variable to forecast LULC changes with \textcolor{revision}{MLP} Markov Neural Networks.

\textcolor{revision}{To the best of our knowledge, there exists no published research  about deep learning models for change forecasting, apart from one notable exception:
\citep{russwurm2020model} employs a recurrent neural network (RNN) to model time series data and forecast low-resolution satellite observations -- e.g., the MODIS NDVI -- in an autoregressive manner. That method does not directly predict urban changes, rather it predicts future satellite observations that may, or may not, enable a subsequent change detection.}

\section{Methodology}

\textcolor{revision}{The question of which pixels in a satellite image will change is ill-posed: change is only defined w.r.t.\ a finite time interval, but a single image does not delimit a time interval (contrary to conventional change detection, where the interval is the time between the two acquisition dates).
Therefore, we must define a time horizon and pose the simpler and more meaningful question of whether or not a change will occur within a given fixed time frame. In a machine learning context this can be accomplished by training only on image pairs with the relevant time window, thus implicitly establishing the forecasting range.}
The limitation to a fixed forecasting range comes with a disadvantage, namely that one can only use a subset of the overall data for training that has the appropriate temporal spacing. In other words, one ignores much of the possibly available data.
To still exploit all samples, we pretrain the backbone on a change detection task (\textit{Stage 1}), where different time spans can be mixed. Subsequently, we finetune the pretrained backbone for the change forecasting tasks (\emph{Stage~2}), \textcolor{revision}{as shown} in Fig.~\ref{fig:graphical_abstract} and described in the following subsections.

\subsection{Stage 1: Change Detection}

In this stage, we train a conventional Siamese network \citep{daudt2018fully,arabi2018optical,yang2021deep} to detect changes on temporally ordered pairs of satellite images. \textcolor{revision}{The network follows the U-Net architecture with ResNet50 encoder \citep{he2016deep}, with shared weights in the two branches and output feature dimension 16 per branch.
The feature maps of the two U-Net branches are concatenated and fed into a \emph{classification head} with two hidden convolutional layers with $(3\times3)$ kernels and depth 16, to obtain the final pixel-wise predictions.}
On the one hand, we can use all the available training pairs at this stage. On the other hand, this type of pretraining already allows the adaptation to the satellite image domain, in contrast to the traditional pretraining that is usually performed on the ImageNet domain.
As a loss function, we use the binary cross-entropy loss.

\subsection{Stage 2: Forecasting}

\textcolor{revision}{For the Forecasting tasks, the pretrained U-Net backbone of \emph{Stage 1} is employed as the base. A new classification head with the same structure as for the change detection task is trained from scratch to perform binary change forecasting and time range forecasting, as described in the subsequent Sections~\ref{sec:mCF} and \ref{sec:mTRF}. 
}

\textcolor{revision}{
\subsubsection{Change Forecasting} \label{sec:mCF}
}

\textcolor{revision}{
The objective of this task is to predict whether a change will occur within a specific, fixed forecasting range. To accomplish this, our model has a single output per pixel, namely a score $\hat{p}_f$ that indicates how likely it is that a building footprint change is expected to occur.
To cover different forecasting ranges, we train nine separate classifiers for ranges of 1, 3, 6, 9, 12, 15, 18, 21, and 24 months, using in each case only the corresponding subset of the training data. We minimize the standard binary cross-entropy loss defined as
    \begin{equation} 
        \mathcal{L}_{binary} = BCE(\hat{p}_f, y_c),
    \end{equation}
where $y_c$ is the binary ``change" / ``no change" label that result from comparing the built-up masks of two time stamps.}

\textcolor{revision}{
\subsubsection{Time Range Forecasting} \label{sec:mTRF} 
}

\textcolor{revision}{
The objective of this model is to classify pixels as belonging to one of two categories: ``early change" (i.e., changes that occur within 1-12 months), or ``late change" (i.e., changes that occur within 12-24 months). We employ a time range forecasting model that has three logit outputs: $q_e$ for ``early", $q_l$ for ``late", and an auxiliary output $q_0$ for the ``no change" class. We use the auxiliary output to obtain additional supervision signals as described below.
}

\textcolor{revision}{
When directly applying multi-class cross-entropy to the problem of classifying pixels into the ``no change", ``early change", and ``late change" categories, the ``no change" class will typically dominate the learning process due to its much higher relative frequency. To address this issue, we split the problem into two sub-problems.
}

\textcolor{revision}{
The first sub-problem is a binary decision between an ``early change" (within 1-12 months) and a ``late change" (within 12-24 months). The predicted score $\hat{p}_e$ is a measure of the likelihood that a given pixel will undergo an ``early change", and is trained with a cross-entropy loss w.r.t.\ the ground truth label $y_e$ (1 for early change, 0 for late change). Note that this loss is only calculated on pixels that exhibit a change as per the GT label.}
    \begin{equation}
        \hat{p}_e = \frac{\exp(q_e)}{\exp(q_e) + \exp(q_l)}
    \end{equation}
    \begin{equation}
        \mathcal{L}_{time} = BCE(\hat{p}_e, y_e).
    \end{equation}

\textcolor{revision}{
The second sub-problem is the 24-month version of the change forecasting task described above in Section~\ref{sec:mCF}, since it aims to classify whether a change within 24 months will occur. The predicted score $\hat{p}_c$ measures the likelihood that a given pixel undergoes a change at all within the maximum time interval. The loss is again a standard cross-entropy between the prediction and the binary ``change" / ``no change" label $y_c$.}
    \begin{equation}
     \hat{p}_c = \frac{\exp(q_e+q_l)}{\exp(q_e+q_l) + \exp(q_0)}
    \end{equation}
    \begin{equation} 
        \mathcal{L}_{binary} = BCE(\hat{p}_c, y_c).
    \end{equation}

\textcolor{revision}{
Finally, we merge the two losses with the mixing weight $\lambda$ to obtain the overall loss for the model. Empirically, $\lambda \approx 10^3$ is a suitable value.}
\textcolor{revision}{
    \begin{equation}
        \mathcal{L} = \mathcal{L}_{time} + \lambda \mathcal{L}_{binary} 
    \end{equation}
}
\textcolor{revision}{
Splitting into two sub-problems with separate losses makes the optimization problem more well-defined. 
Dividing the task into two sub-problems each with its own losses, results in a better formulated optimization problem. The time range forecasting term can only be calculated on a small portion of the available pixels (the ones that indeed exhibit a change as per the GT), but it provides an almost perfect class balance between ``early" and ``late" changes. On the other hand, the loss provided by the binary change label is calculated from the complete set of pixels but suffers from a class imbalance issue.
Overall, the combination of the loss terms improves the performance of the resulting model.
}

\textcolor{revision}{\subsection{Label Imbalance}}

\textcolor{revision}{
Change detection, and by extension also change forecasting, typically suffers from severe label imbalance. We thus employ two balancing strategies, in both stages of the training workflow.}

First, we oversample change examples. Image patches $i$ are sampled with probability  
\begin{equation}
    p_{i} = \frac{a + N_{i}}{\sum^M_k a + N_{k}},
\end{equation}
where \textcolor{revision}{$N_i$} is the number of changed pixels per patch i; $M$ is the number of samples; and $a$ is a distribution smoothing constant, set to $a \approx 50$, a value that was found empirically.  

\textcolor{revision}{Second,} we noticed that thresholding the scores at 0.5 \textcolor{revision}{(respectively at 0.33 for the 3-class time range forecasting)}  results in a poor precision-recall trade-off with a heavy bias towards the majority  class ``no change''(we further elaborate on this effect in Section~\ref{sec:results}).
\textcolor{revision}{To counteract the bias, we determine the threshold in a data-driven manner: separately for every training batch, we find the threshold that maximizes the F1 score. To reduce the effect of stochasticity in the training mini-batches, we compute the final threshold value as a moving average over the last 500 training batches.}
Empirically, the approximation is very good: the discrepancy between the threshold estimated in this manner and the \textcolor{revision}{oracle threshold determined from the test set} is vanishingly small.\\

\subsection{Implementation Details}

Our method is implemented in PyTorch~\citep{NEURIPS2019_9015} and is publicly available\footnote{https://github.com/nandometzger/ChangePrediction}.
To train the model we use the Adam optimizer \citep{kingma2014adam} with default parameters and a base learning rate of $10^{-4}$. We augment the samples by random cropping, small affine transformations, mirroring, and color jittering. We use a batch size of 16, except for the largest forecasting ranges of 21 and 24, where we found batch size 4 to perform best, likely as a consequence of the small size of the corresponding data subset.

For all models in \emph{Stage 2}, we first freeze the backbone and train the head for 5,000 \textcolor{revision}{batch} iterations, then we reduce the learning rate by a factor of 10 to $10^{-5}$ and train the entire model end-to-end.

\section{Data and Experimental Setup}

\begin{figure}
    \centering
    \includegraphics[width=0.45\textwidth]{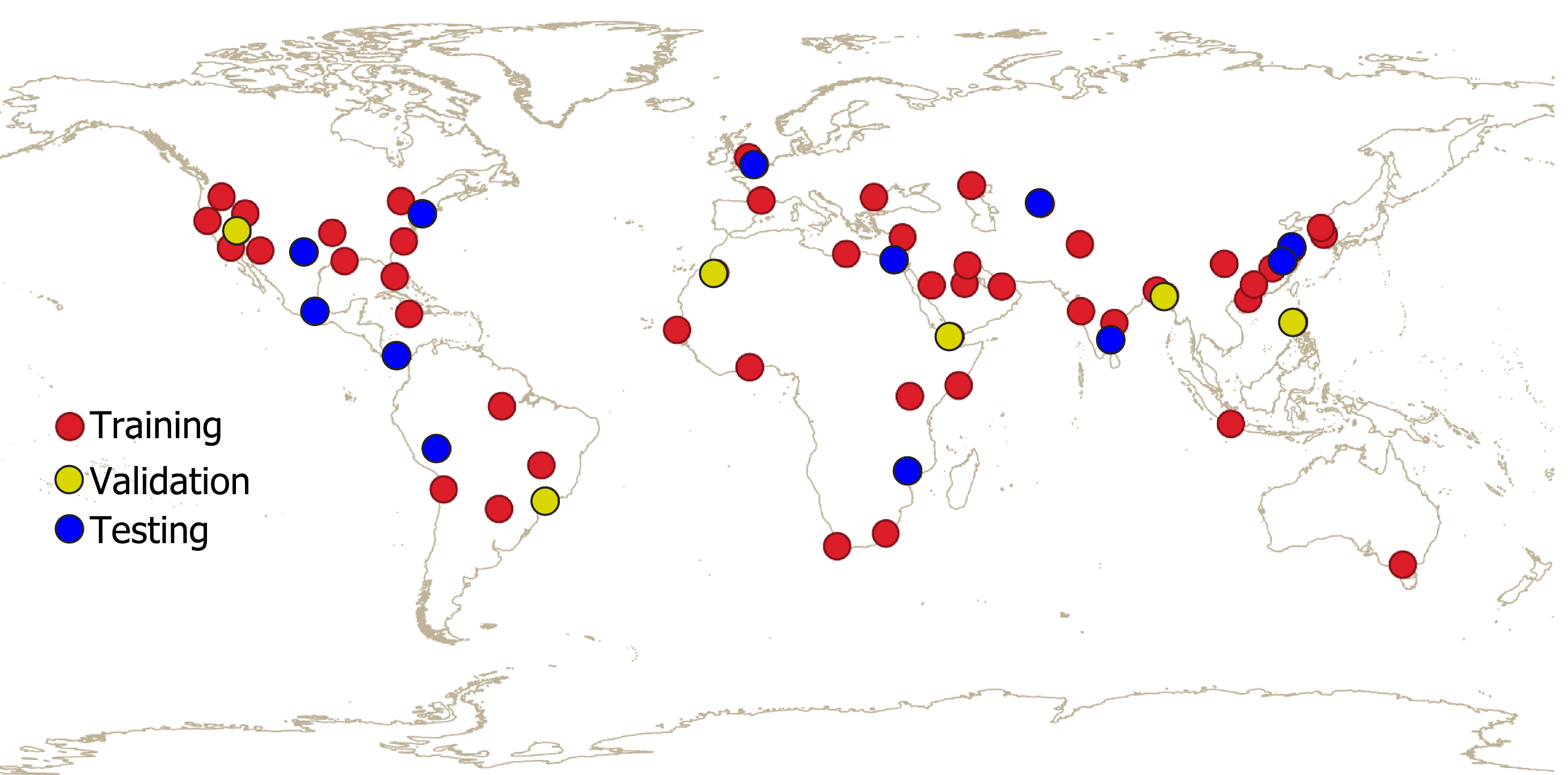}
    \caption{Spatial distribution of the SpaceNet7 dataset \textcolor{revision}{and train/validation/test split.}}
    \label{fig:SpaceNet7map}
\end{figure}

To validate our method, we make use of the SpaceNet7 dataset. It has been published by \citet{van2021multi} and \textcolor{revision}{was} created for the building tracking competition featured at NeurIPS 2020. The dataset consists of 60 globally distributed labeled locations (see Fig.~\ref{fig:SpaceNet7map}), each containing a series of 24 Planet Labs RGB satellite image mosaics of 4x4 km\textsuperscript{2}, with consecutive mosaics, acquired one month apart. The ground sampling distance of the images is 4 m, and the total covered area is 960 km\textsuperscript{2}. The dataset also contains a set of manually labeled building footprints, where each image of the time series is labeled individually.

\begin{figure} [t]
    \vspace{0.3cm}
    \centering
    \begin{tikzpicture}
    \begin{axis}[%
    grid=major,grid style={dashed},
    width=0.3\textwidth,
    height=2.25in,
    scale only axis,
    xmin=0,
    xmax=25,
    separate axis lines,
    every outer y axis line/.append style={black},
    every y tick label/.append style={font=\color{black}},
    ymin=0,
    ymax=25000,
    ylabel      ={\ref{displacementplot} Number of patches (224$\times$224)},
    xlabel={Forecasting range [months]},
    xtick       ={1,3,6,9,12,15,18,21,24},
    xticklabels ={1,3,6,9,12,15,18,21,24},
    ]
    \addplot [
    color=red,
    solid,
    mark=x
    ]
    coordinates {
    	(1,22600)
    	(3,20500)
        (6,17800) 
        (9,15000) 
        (12,12400) 
        (15,9500) 
        (18,6700) 
        (21,4100) 
        (24,1200)}; 
    
    \label{displacementplot}
    \end{axis}
    
    \begin{axis}[%
    width=0.3\textwidth,
    height=2.25in,
    scale only axis,
    xmin=0,
    xmax=25,
    ymin=0,
    ymax=1.375,
    ytick={   0, 0.25, 0.5, 0.75, 1.0, 1.25},
    yticklabels ={    0$\%$, 0.25$\%$, 0.5$\%$, 0.75$\%$,  1.0$\%$, 1.25$\%$},
    every outer y axis line/.append style={black},
    every y tick label/.append style={font=\color{black}},
    ylabel={\ref{areaplot} Percentage of \textcolor{revision}{changed} pixels},
    xtick       ={1,3,6,9,12,15,18,21,24},
    xticklabels ={1,3,6,9,12,15,18,21,24},
    axis x line*=bottom,
    axis y line*=right,
    xtick={},
    xticklabels={},
    ]
    \addplot [
    color=blue,
    solid,
    mark=x
    ]
    coordinates {
    	(1,0.0418526785714)
    	(3,0.1215720663265)
        (6,0.2431441326531) 
        (9,0.3687021683673) 
        (12,0.4902742346939) 
        (15,0.6218112244898) 
        (18,0.7453762755102) 
        (21,0.8510044642857) 
        (24,1.1459661989796)}
    ;\label{areaplot}

    \end{axis}
    \end{tikzpicture}%
    \caption[Size and imbalance of SpaceNet7 dataset.]{Comparison between the number of available samples and the balance of the dataset depending on the size of the chosen forecasting range.}
    \label{fig:size_balance}
\end{figure}
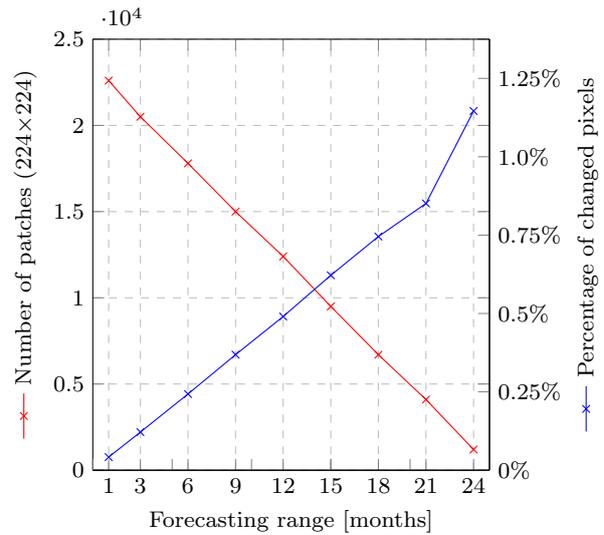

For this work, we derive a dataset that considers image pairs from the same locations but at different time steps and \textcolor{revision}{obtain change masks by subtracting the corresponding built-up area masks from each other.}
\textcolor{revision}{We note that, for simplicity, we have omitted cases where buildings have been removed. The reason is that our goal was a dataset that showcases urban development, as opposed to the opposite scenario where buildings are removed for good. The latter is far less common and involves completely different visual cues. During visual inspection, we found that many of the apparent destruction labels were actually caused by misalignment errors between the manually digitized footprints at different times, rather than actual building destruction.
}
Using this approach, we obtain about 16,000 unique image pairs, which we further split into 264,000 non-overlapping pairs of training patches of size 224$\times$224 pixels.
However, for the task of change forecasting, one implicitly defines the forecasting range by the choice of training sub-datasets that have a consistent forecasting range.
For example, when we want to forecast for the smallest possible range (i.e., one month), the subset of pairs that are one month apart contains 5,000 samples, whereas, for the largest forecasting range of 24 months, we only obtain 200 samples.
Moreover, the dataset also exhibits a severe label imbalance, which is quite common for change detection datasets\textcolor{revision}{~\citep{daudt2019multitask}}.
The average fraction of changed pixels across all samples amounts to 0.3\%, and only one in seven patches have \textgreater0.5\% positive labels. 
Fig.~\ref{fig:size_balance} summarizes the size and imbalance of the sub-datasets.

\textcolor{revision}{To ensure a representative and balanced evaluation, we split the dataset stratified with respect to the individual continents, resulting in} a training set with 42 locations (70\%), a validation set for parameter tuning with 6 locations (10\%), and a test set to calculate the final metrics with 12 locations (20\%). \textcolor{revision}{The exact division of the geographical areas is illustrated in Fig.~\ref{fig:SpaceNet7map}}

\vspace{1em}
\section{Results and Discussion} \label{sec:results}

\subsection{Change Forecasting} \label{res:change_fore}

\begin{figure} [t]
\centering
\resizebox{0.95\columnwidth}{!}{
\begin{tikzpicture}
    \begin{axis}[    xlabel=Prediction Range (months),    ylabel=F1 score (\%),    xmin=0, xmax=24,    ymin=0, ymax=55,    xtick={1, 6, 12, 18, 24},    ytick={0,5,10,15, 20, 25,30,35,40,45, 50},    legend pos=north west,    ymajorgrids=true,    grid style=dashed,] 
    \addplot[    color=orange,    mark=*,    ]
        coordinates {
                    (	1	,	15.33	)
                    (	3	,	22.69	)
                    (	6	,	28.96	)
                    (	9	,	31.14	)
                    (	12	,	33.1	)
                    (	15	,	34.35	)
                    (	18	,	35.32	)
                    (	21	,	38.22	)
                    (	24	,	41.67	)
        };
    \addplot[    color=brown,    mark=*,    ]
        coordinates {
                    (	1	,	11.71	)
                    (	3	,	19.56	)
                    (	6	,	26.47	)
                    (	9	,	28.43	)
                    (	12	,	30.01	)
                    (	15	,	32.51	)
                    (	18	,	34.12	)
                    (	21	,	36.54	)
                    (	24	,	41.36	)
        }; 
    \addplot[    color=blue,    mark=square,    ]
        coordinates {
            	(1,   7.977)
            	(3,   6.263)
            	(6,   10.75)
            	(9,   10.3)
            	(12,  12.41)
            	(15,   9.522)
            	(18,   11.32)
            	(21,   9.941)
            	(24,   15.63)
        };
    \addplot[    color=red,    mark=square,    ]
        coordinates {
                    (	1	,	1.129	)
                    (	3	,	3.954	)
                    (	6	,	7.488	)
                    (	9	,	6.422	)
                    (	12	,	9.908	)
                    (	15	,	8.315	)
                    (	18	,	8.943	)
                    (	21	,	7.866	)
                    (	24	,	13.36	)
        };
    
    \legend{Detection (ImageNet), Detection (vanilla), Forecasting (ours), Forecasting (ImageNet)}
    \end{axis}
\end{tikzpicture}
    }
\caption{Performance of the change detection and change forecasting \textcolor{revision}{models in terms of F1 score of the foreground class}.}
\label{fig:results}
\end{figure}
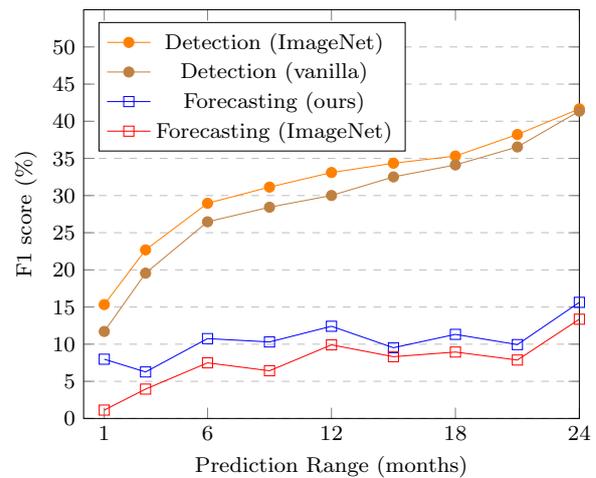

In the comparison of Fig.~\ref{fig:results}, we plot the F1 scores w.r.t.\ the forecasting range of our method (denoted as ``Forecasting (ours)") and compare it to a baseline pretrained in single-image mode on ImageNet (``Forecasting (ImageNet)"), rather than on the satellite image change detection task. Moreover, we also provide the change detection performance trained from scratch (``Detection (vanilla)") and pretrained on ImageNet (``Detection (ImageNet)"), as an upper bound for what the forecasting model can be expected to achieve:
it would be unreasonable to expect the model to perform better in forecasting changes using a single image than in detecting changes when a second image is also available.
Note that there is only one change detection model for all ranges, while there are separately fine-tuned change forecasting models for different prediction ranges.

Our model exhibits a consistent gain of 2-3 percentage points over the baseline approach.
The advantage is most pronounced for the 1-month range, where our model outperforms the baseline with 8.0\% vs.\ 1.1\%.
This is a particularly challenging forecasting range for learned models because it suffers from the most severe imbalance in the labels: there are very few positive pixels, as within 1 month only few construction projects proceed to a point where \textcolor{revision}{a building is clearly recognizable}. It appears that a better initialization of the network weights \textcolor{revision}{yields} higher robustness against such extreme label distributions, such that forecasting performance actually matches the change detection model.
With the increasing prediction range, the changes get more substantial\textcolor{revision}{,} meaning that the detection task gets easier, whereas forecasting in the absence of a second image remains equally hard.
Consequently, the change detector benefits from less imbalanced labels and significantly improves as the temporal range grows, while the performance of change forecasting remains at a respectable 10\% across most of the range, increasing to 15\% for the 24-month range.

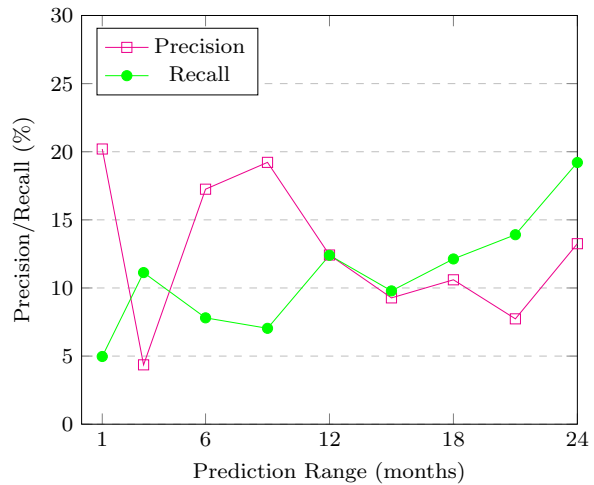
\begin{figure} [t]
\centering
\resizebox{0.95\columnwidth}{!}{
\begin{tikzpicture}
    \begin{axis}[    xlabel=Prediction Range (months),    ylabel=Precision/Recall (\%),    xmin=0, xmax=24,    ymin=0, ymax=30,    xtick={1, 6, 12, 18, 24},    ytick={0,5,10,15, 20, 25,30,35,40,45},    legend pos=north west,    ymajorgrids=true,    grid style=dashed,]
    \addplot[    color=magenta,    mark=square,    ]
        coordinates {
                    (	1	,	20.2	)
                    (	3	,	4.358	)
                    (	6	,	17.25	)
                    (	9	,	19.22	)
                    (	12	,	12.42	)
                    (	15	,	9.272	)
                    (	18	,	10.6	)
                    (	21	,	7.734	)
                    (	24	,	13.25	)
        };
    \addplot[    color=green,    mark=*,    ]
        coordinates {
                    (	1	,	4.97	)
                    (	3	,	11.13	)
                    (	6	,	7.807	)
                    (	9	,	7.038	)
                    (	12	,	12.4	)
                    (	15	,	9.786	)
                    (	18	,	12.13	)
                    (	21	,	13.91	)
                    (	24	,	19.21	)
        };
    \legend{Precision, Recall}
    \end{axis}
\end{tikzpicture}
}
\caption{Precision and recall \textcolor{revision}{of the foreground class} of our change forecasting model w.r.t.\ the prediction range. \textcolor{revision}{The shown numbers correspond to the F1 score of the ``ours" model in Fig.~4.} }
\label{fig:precision_recall}
\end{figure}

\begin{figure}
    \centering
    \includegraphics[width=0.95\columnwidth]{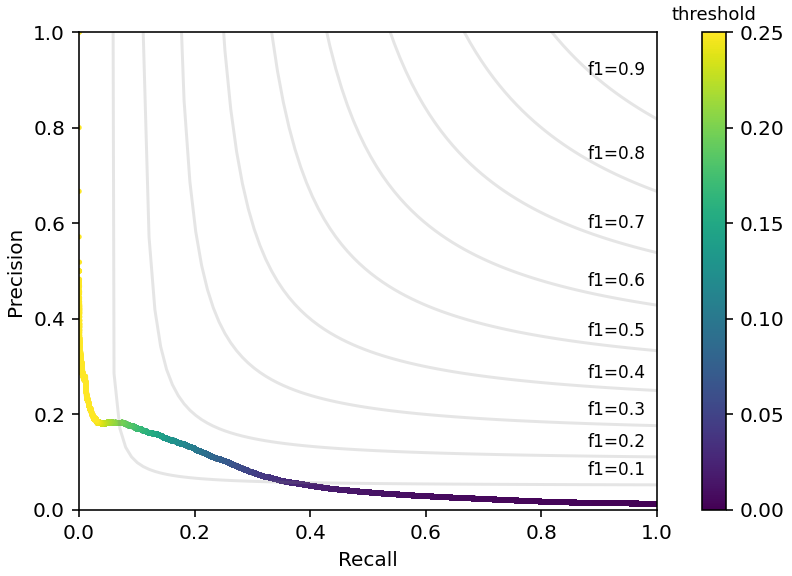}
    \caption{Precision-Recall curve for the \textcolor{revision}{change forecast model with time horizon 24 months}.}
    \label{fig:PRcurve24months}
\end{figure}

\begin{figure*} [tb]
    \centering
    \includegraphics[width=0.95\textwidth]{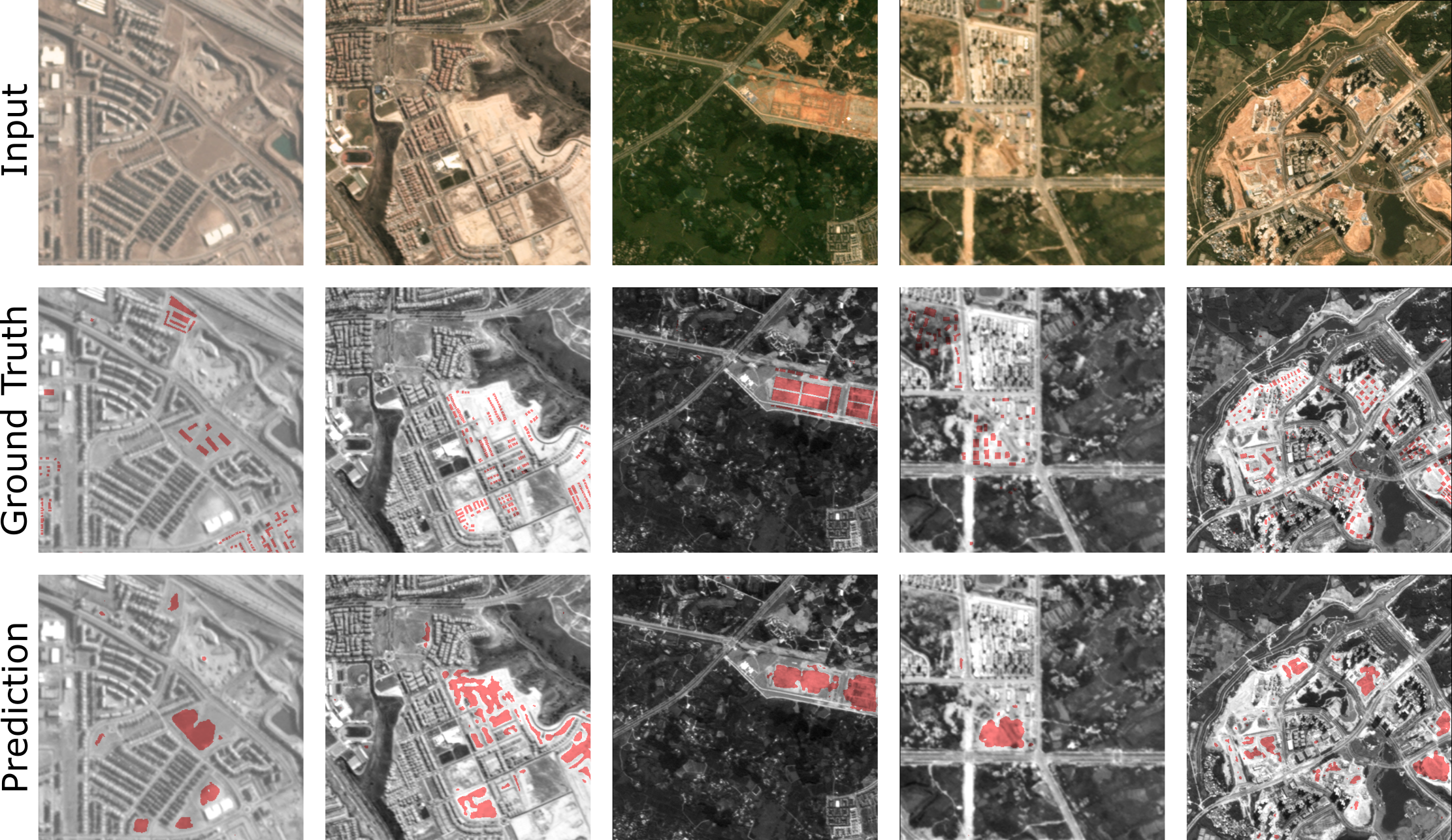}
    \caption{Comparison of model prediction for the 24-month prediction range and the corresponding true change, overlaid on the panchromatic input image. Even with a single input image, the model finds many locations of building changes in the 24 months prediction range. }
    \label{fig:viz_pos}
\end{figure*}

\begin{figure} [tb!]
    \centering
    \includegraphics[width=0.45\textwidth]{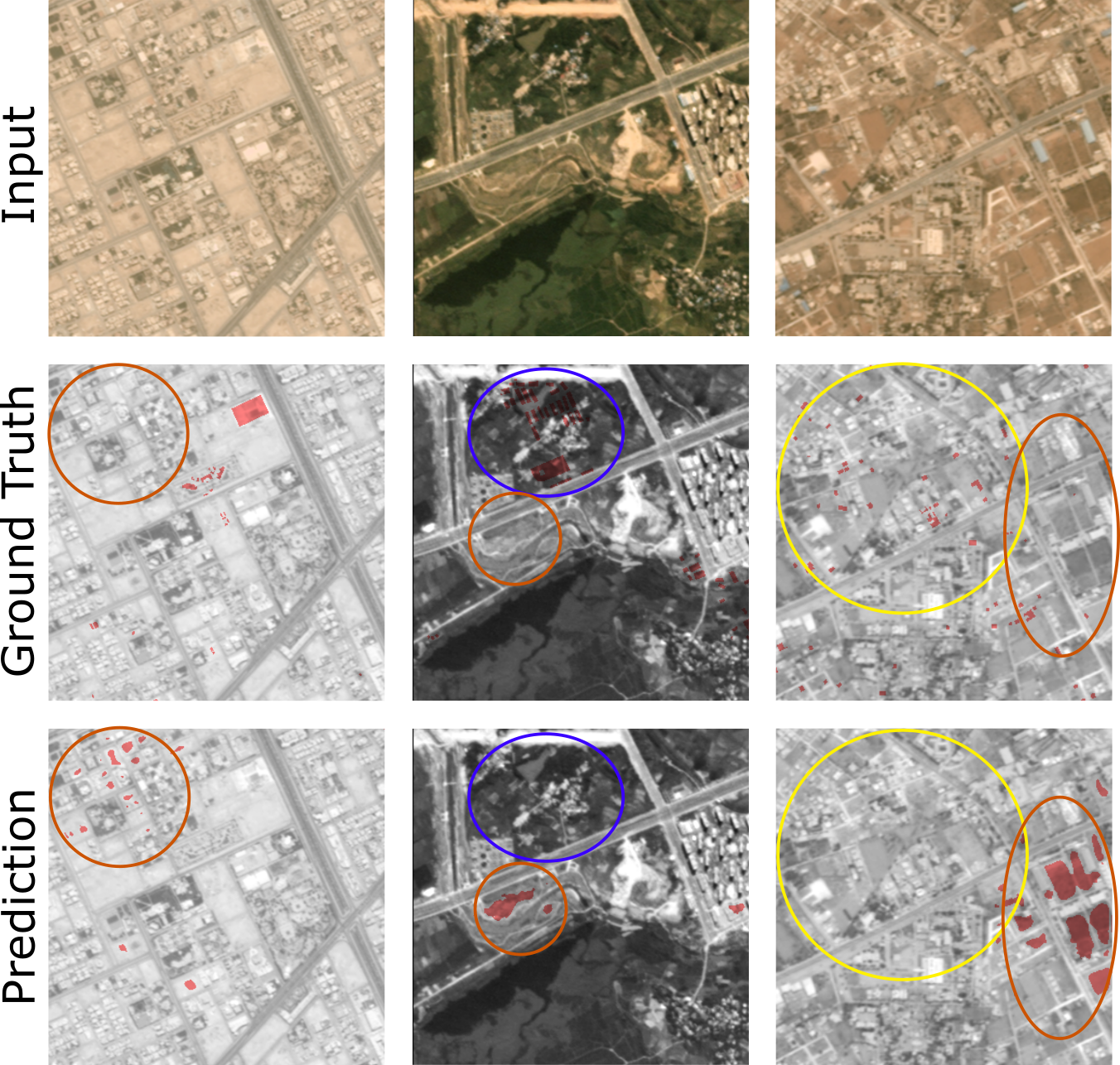}
    \caption{Typical failure cases: A significant amount of false positives are caused by areas that resemble construction sites \textcolor{revision}{ (orange markers).}
    \textcolor{revision}{Understandably, our model struggles when there is no indication of imminent construction at all -- producing false negatives (blue marker). Moreover, our model fails to detect small, but distributed changes (yellow markers). Samples are from the 24 months prediction range.}   }
    \label{fig:viz_neg}
\end{figure}

In Fig.~\ref{fig:precision_recall}, we show the precision and recall scores that correspond to the F1 scores previously discussed.
\textcolor{revision}{
The trade-off between precision and recall for each interval is influenced by the nature of individual sub-datasets and the classification threshold. We hypothesize that the observed trend of higher precision for the first 12 months and higher recall for the second 12 months may be attributed to the reduced class imbalance in longer forecasting ranges. Under the assumption that the models focuses to a significant degree on detecting construction sites, it seems reasonable that the recall would gradually increase. This is because construction sites foreshadow urban changes. With increasing forecasting range these sites are more likely to reach a point where new buildings have been erected, leading to more positive labels and thus higher recall.}

Moreover, we provide the breakdown of the precision-recall trad\textcolor{revision}{e}-off curve in Fig.~\ref{fig:PRcurve24months} for the classifier fine-tuned to the 24 months prediction range. 
The curve exhibits a bias toward favoring recall over precision -- meaning, a significant amount of recall must be relinquished to see an increase in precision larger than 20\%.
This may indicate that in some cases the image evidence is sufficient to anticipate an imminent change, but not to localize it. Intuitively, this seems to make sense, as grading and earthworks in early stages do reveal the intention to construct, but not the location of the individual buildings within the plot.

Fig.~\ref{fig:viz_pos} illustrates successful predictions by our model. \textcolor{revision}{The model tends to identify the rough location of future changes correctly, mostly on the basis of detecting construction sites. 
The shape of the model predictions generally does not align exactly with the ground truth. This is not surprising given the inherent ambiguity of the task -- from the early earthworks and preparations it is not possible to determine the precise outlines of the future buildings. Moreover, CNNs by construction tend to produce blurry outputs in the presence of uncertainty. Fortunately, knowing the location and the approximate extent of the land cover change is sufficient for many downstream tasks.
} 

In Fig.~\ref{fig:viz_neg} we display typical failure cases, which further help to understand which visual cues the model relies on for its predictions.
It is apparent that the model anticipates new buildings at early-stage construction sites, but it seems to also have acquired a rudimentary understanding of urban development and sprawl, as it tends to predict the construction of new buildings in cluttered or empty wastelands that lie in the vicinity of existing buildings.

\subsection{Time Range Forecasting}

\textcolor{revision}{
To isolate the time range prediction from the binary change forecasting, we restrict the following evaluation to pixels that do exhibit a change according to the ground truth labels.}
\textcolor{revision}{
    In Table~\ref{tab:timerange_main}, we present the direct comparison of our pretraining method and the standard ImageNet pretraining in terms of accuracy (Acc) and average F1 score (aF1). The results show that our custom pretraining approach indeed improves performance by 3\% in both accuracy and F1 score, further supporting the effectiveness of our proposed methodology.
}

\begin{figure} [t]
    \centering
    \includegraphics[width=0.45\textwidth]{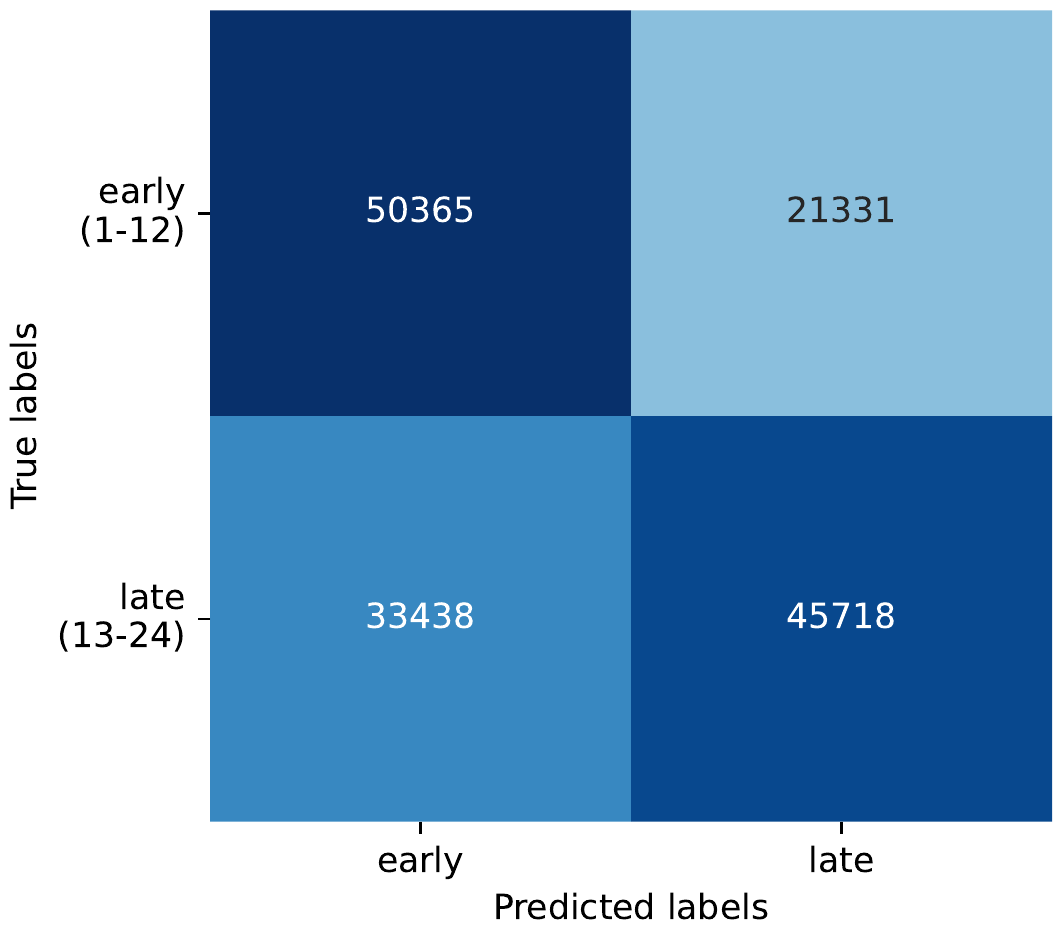}
    \caption{The confusion matrix of our approach for time range forecasting with classes early changes (1-12 months) and late changes (13-24 months) \textcolor{revision}{which equates to an accuracy of 64.0\%. The precision for the ``early" class amounts to 60.0\%, while the one for the ``late" class amounts to 68.2\%.}
    }
    \label{fig:cm}
\end{figure}

We present the confusion matrix for time range forecasting in Fig.~\ref{fig:cm}.
It shows that with our setup it is in principle also possible to classify future change events into a group that will happen sooner and another one that will only happen later. \textcolor{revision}{
We note that the ``early" class exhibits a precision of 60.0\%, while the precision of the ``late" class amounts to 68.2\%, suggesting that the later changes might be easier to detect than the earlier ones.
}
\textcolor{revision}{
    Table~\ref{tab:timerange_ablation} presents a comparison of our approach to models specifically trained for time range forecasting and change forecasting, respectively. For the time range forecasting, we use the same evaluation procedure as before, e.g. restricting the evaluation to pixels that do exhibit a change according to the ground truth labels.
    Moreover, we provide the accuracy and average F1 score over the ``early" and ``late" changes. Additionally, we display the F1 score, precision (Pre), and recall (Rec) metrics for the ``change" class from the change forecasting task. The results indicate that our multi-task approach is beneficial for the time range task, but not for the binary change forecasting, as it trades off recall for precision.
}

\begin{table}[t]
\centering 
\textcolor{revision}{
\begin{tabular}{@{}lcc@{}} 
                              & Acc {[}\%{]}   & aF1 {[}\%{]}     \\ \midrule
ImageNet pretraining          & 61.1           & 60.1          \\
Ours                          & \textbf{64.0}  & \textbf{63.7}    \\
\bottomrule
\end{tabular}%
}
\caption{\textcolor{revision}{Comparison of our pretraining approach against the ImageNet pretraining approach on the time range forecasting task.  } }
\label{tab:timerange_main}
\end{table}

\textcolor{revision}{
    Additionally, we present an empirical examination of the mixing weight parameter $\lambda$ in Table~\ref{tab:mix_weight}. The analysis shows that the model performs best at 1000, but breaks for values $<$10 and $>$1000.
}

\begin{table}[t]
\centering
\resizebox{0.475\textwidth}{!}{%
\textcolor{revision}{
\begin{tabular}{@{}lccccc@{}}
\cmidrule(l){2-6}
              & \multicolumn{2}{c}{Time Range}   & \multicolumn{3}{c}{Change Forecasting}      \\ \cmidrule(l){2-6} 
              & Acc {[}\%{]} & aF1 {[}\%{]} & F1 {[}\%{]} & Pre {[}\%{]} & Rec {[}\%{]} \\ \midrule
Range only & 58.6    & 60.0 & - & - & -            \\
Forecasting only & -   & -  & \textbf{15.2} & \textbf{25.1} & 10.9 \\
Both           & \textbf{64.0} & \textbf{63.7}  & 5.3 & 2.8 & \textbf{54.7}         \\ \bottomrule
\end{tabular}%
}
}
\caption{\textcolor{revision}{Combining both loss functions boosts the time range forecasting task, while it deteriorates the performance in the change forecasting task.} }
\label{tab:timerange_ablation}
\end{table}

\begin{table}[t]
\centering
\textcolor{revision}{
\begin{tabular}{@{}lcc@{}} 
$\lambda$ & \multicolumn{1}{c}{Acc {[}\%{]}} & \multicolumn{1}{c}{aF1 {[}\%{]}} \\ \midrule
0.1                  & 53.1          & 47.3             \\
1                    & 55.9          & 55.9             \\
10                   & 60.9          & 60.8             \\
100                  & 61.1          & 60.4             \\
1000 (ours)                & \textbf{64.0} & \textbf{63.7}    \\
10000                & 54.9          & 54.8               \\ \bottomrule
\end{tabular}
}
\caption{ \textcolor{revision}{Analysis of the mixing weight parameter $\lambda$.
}}
\label{tab:mix_weight}
\end{table}

\section{Conclusion}

The main contribution of this work is a model to forecast where new buildings are likely to appear in the near future. Our goal has been to \textcolor{revision}{present a contribution to} this little-explored topic in \textcolor{revision}{the} light of modern deep learning technology.
Besides setting a first baseline for change forecasting with deep convolutional networks, we have designed a 2-stage transfer learning procedure that employs change detection \textcolor{revision}{from paired images} as a proxy task for learning features tailored to the analysis of high-resolution satellite images of urban and periurban regions.
We have shown that such a pretraining improves change forecasting across a range of time horizons and that it is particularly helpful for a short horizon of 1 month, where the imbalance between unchanged and changed areas is particularly extreme.
Moreover, we have also shown that it is possible, to some degree, to forecast how far into the future a change is going to happen.

Clearly, the presented approach does not perfectly solve the problem, mainly due to the fact that forecasting a future construction event from a single image is an ill-posed and very challenging problem.
On the one hand, while there obviously are visual cues for future construction activity, none of these cues are guaranteed and unambiguous. For instance, earthworks may point at future construction, but they can take place for other reasons, e.g., landscaping. Furthermore, if a new estate will be constructed, there may already be signs like access roads or earthworks 2 years before, but there could also still be grassland or even agricultural fields, etc.
Besides these conceptual limits, there are also technical challenges like the difficulty to obtain a large enough foreground set of comparatively infrequent events, such as new construction and the associated imbalance of the available labels.

We consider our work as a first attempt and hope that it may encourage further research and development toward more powerful and sophisticated forecasting methods.
While we have tried to pioneer the use of deep learning for this type of forecasting, our standard convolutional design only scratches the surface of what is possible. Especially if multiple pre-event images are available, it would seem natural to explicitly model the temporal evolution of urban development -- for instance with recurrent or attention-based architectures -- to better exploit the temporal characteristic of the data.
\textcolor{revision}{One way to overcome the scarcity of data may also be to synthetically introduce changes in images if one manages to bridge the domain gap between real and synthetic examples.}
Going for longer-term forecasts significantly beyond the next 2 years is a formidable challenge in terms of both data and methods, but would open up opportunities for a whole new set of applications such as long-term infrastructure planning tasks.

\section*{Declarations}

\paragraph{Acknowledgements}
Special thanks go to Stefano D'Aronco for his many valuable technical suggestions. Moreover, we thank Corentin Perret-Gentil and Philipp Krüsi for their practical input and Daedalean AI for the collaboration.

\paragraph{Competing interests}
The authors declare no competing interests.

\paragraph{Availability of data and material} 
The SpaceNet7 dataset is publicly available \citep{van2021multi}.

\paragraph{Code availability} 
The code is available at \url{https://github.com/nandometzger/ChangePrediction}.

\paragraph{Authors' contributions}
KS and JDW conceived the study. MÖT, RCD provided technical inputs. KS, RCD proofread the manuscript. NM implemented the methods. conducted the experiments and created the raw version of the manuscript.
 
\bibliographystyle{spbasic}      
\bibliography{bibliography}   

\end{document}